\documentclass[letterpaper]{article} 
\usepackage{aaai2026}  
\usepackage{times}  
\usepackage{helvet}  
\usepackage{courier}  
\usepackage[hyphens]{url}  
\usepackage{graphicx} 
\usepackage{natbib}  
\usepackage{caption} 
\usepackage{algorithm}
\usepackage{algorithmic}
\usepackage{booktabs}
\usepackage{multirow} 
\usepackage{color}
\usepackage{amsmath}
\usepackage{amssymb}
\usepackage{pifont}
\usepackage{colortbl}
\usepackage{xcolor}
\usepackage[table]{xcolor}
\usepackage{arydshln}
\usepackage{times}
\usepackage{helvet}
\usepackage{courier}
\usepackage{bibentry}
\usepackage{newfloat}
\usepackage{listings}
\DeclareCaptionStyle{ruled}{labelfont=normalfont,labelsep=colon,strut=off} 
\floatstyle{ruled}
\newfloat{listing}{tb}{lst}{}
\floatname{listing}{Listing}
\title{D\textsuperscript{2}-VPR: A Parameter-efficient Visual-foundation-model-based Visual Place Recognition Method via Knowledge Distillation and Deformable Aggregation}
\author{
    Zheyuan Zhang\textsuperscript{\rm 1, \rm 2}, 
    Jiwei Zhang\textsuperscript{\rm 1, \rm 2}, Boyu Zhou\textsuperscript{\rm 1, \rm 2}, Linzhimeng Duan\textsuperscript{\rm 1, \rm 2}, Hong Chen\textsuperscript{\rm 1, \rm 2}\thanks{Hong Chen is the corresponding author.}
}
\affiliations{
    \textsuperscript{\rm 1}School of Computer Science, Beijing University of Posts and Telecommunications\\

     \textsuperscript{\rm 2}Key Laboratory of Interactive Technology and Experience System, Ministry of Culture and Tourism,\\Beijing University of Posts and Telecommunications\\
     \{zzy1998, chenhong76\}@bupt.edu.cn

}

\usepackage{bibentry}

\begin{document}

\maketitle

\begin{abstract}
Visual Place Recognition (VPR) aims to determine the geographic location of a query image by retrieving its most visually similar counterpart from a geo-tagged reference database. Recently, the emergence of the powerful visual foundation model, DINOv2, trained in a self-supervised manner on massive datasets, has significantly improved VPR performance. This improvement stems from DINOv2’s exceptional feature generalization capabilities but is often accompanied by increased model complexity and computational overhead that impede deployment on resource-constrained devices. To address this challenge, we propose $D^{2}$-VPR, a $D$istillation- and $D$eformable-based framework that retains the strong feature extraction capabilities of visual foundation models while significantly reducing model parameters and achieving a more favorable performance-efficiency trade-off. Specifically, first, we employ a two-stage training strategy that integrates knowledge distillation and fine-tuning. Additionally, we introduce a Distillation Recovery Module (DRM) to better align the feature spaces between the teacher and student models, thereby minimizing knowledge transfer losses to the greatest extent possible. Second, we design a Top-Down-attention-based Deformable Aggregator (TDDA) that leverages global semantic features to dynamically and adaptively adjust the Regions of Interest (ROI) used for aggregation, thereby improving adaptability to irregular structures. Extensive experiments demonstrate that our method achieves competitive performance compared to state-of-the-art approaches. Meanwhile, it reduces the parameter count by approximately 64.2\% (compared to CricaVPR).
\end{abstract}

\begin{links}
    \link{Code}{https://github.com/tony19980810/D2VPR}
\end{links}

\section{Introduction}

\begin{figure}[t]
    \centering
    \includegraphics[width=0.9\linewidth]{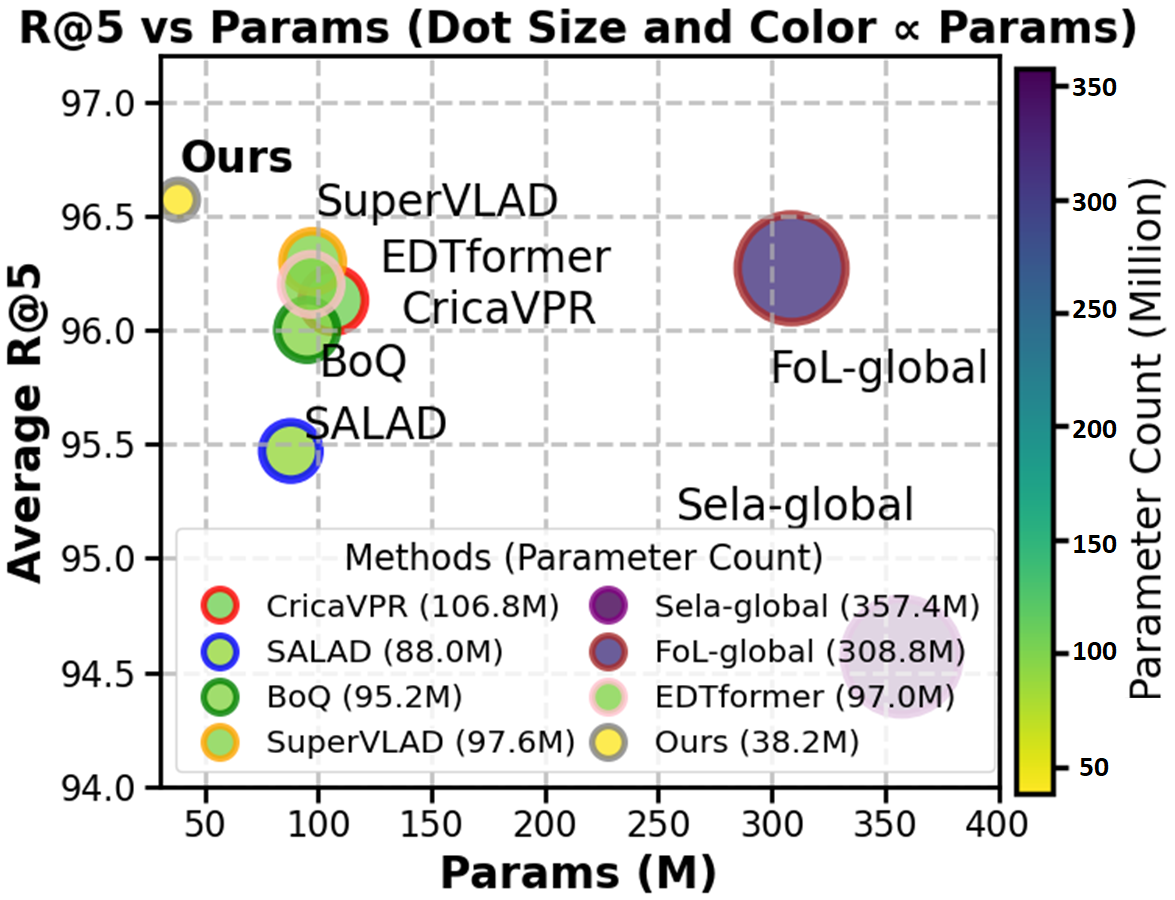}
    \caption{The comparison of average R@5 against parameter count on Pitts30k, MSLS-val, and SPED (with image size of 224×224) demonstrates that our model achieves competitive performance despite significantly reduced parameter count, striking an effective trade-off.}
\label{intro}
\vspace{-0.3cm}
\end{figure}

Visual Place Recognition (VPR) is to identify the location where a query image is captured by finding the most visually similar image in a geo-tagged reference database~\cite{chen2017deep, netvlad}. It serves as a cornerstone for numerous real-world applications, including autonomous robot navigation~\cite{lowry2015visual}, augmented reality~\cite{ventura2014global}, and location-based services~\cite{sarlin2019coarse}. VPR offers advantages over other sensing modalities (e.g., LiDAR or RADAR) such as lower cost, easier deployment and passive data acquisition, making it an attractive option for many applications~\cite{10475537}. However, VPR still faces major challenges from long-term appearance variations (e.g., due to seasonal, weather, or illumination changes), perceptual aliasing (where distinct places appear deceptively similar) and viewpoint shifts~\cite{torii201524}.

Recently, with further breakthroughs in foundational vision models, VPR approaches~\cite{lu2024supervlad, ali2024boq, salad, crica, sela, EDTformer, wang2025focus} using DINOv2~\cite{oquab2023dinov2} as the backbone, to some extent, have mitigated the aforementioned challenges. DINOv2, a self-supervised vision transformer pretrained on 142 million diverse web images (spanning varied lighting, seasons, and viewpoints) and filtered via embedding clustering for diversity and quality, leverages large-scale exposure for feature generalization. Combined with a self-supervised teacher-student distillation strategy, it produces patch-level features robust to long-term appearance changes, perceptual aliasing, and viewpoint shifts. Yet, while yielding more generalizable features, the massive parameter counts of visual foundation models restrict the deployment and application of methods built on them to resource-constrained devices. For instance, LPS-VPR~\cite{nie2024efficient}—a method using a convolutional neural network (CNN) architecture—has an overall parameter count of 30M. In contrast, existing DINOv2-based VPR methods mostly rely on DINOv2-base (with 86M parameters for the backbone alone) or DINOv2-large (with 300M parameters for the backbone alone).

To preserve the strong feature representation of vision foundation models while substantially reducing parameter count, a natural and direct strategy is to adopt the smaller visual foundation model, e.g., DINOv2‑small (with 22M parameters for the backbone alone). However, when replicating the training pipeline of CricaVPR~\cite{crica} using DINOv2‑small as the backbone, our experiments (see Ablation Study) show a pronounced drop in performance compared to using DINOv2‑base. This explains why current methods have not selected the smaller variant as the backbone. To overcome this limitation, we still retain the more lightweight DINOv2-small as our backbone. However, instead of directly conducting training related to the VPR task, we adopt a two-stage training strategy—knowledge distillation followed by fine-tuning—which dramatically reduces the parameter count while preserving the rich representational power of the visual foundation model. To minimize knowledge loss during transfer, we introduce a distillation recovery module that aligns teacher and student features through the fusion of shallow and deep representations. Furthermore, to enhance the capacity of spatial-pooling-based aggregators (e.g., CricaVPR's) to represent irregular geometric structures, we design a flexible deformable aggregator that dynamically adapts pooling regions to better capture complex spatial relationships. Our deformable aggregator, inspired by neural top-down attention~\cite{lou2025overlock}, combines semantic and global features to dynamically deform pooling regions so they precisely fit and emphasize irregular, key local areas. Afterwards, these focused local representations are fed back to reinforce the global features, creating a bidirectional interaction that enables the network to both guide its attention based on semantics and refine its understanding of overall context. Incorporating the improvements mentioned above, our method demonstrates strong competitiveness in both parameter efficiency and performance. While significantly reducing the number of parameters, it retains the robust feature representation capabilities of visual foundation models, showing competitive performance compared to existing state-of-the-art (SOTA) models on popular benchmarks, as illustrated in Figure \ref{intro}.

To summarize, our work makes the following contributions: \textbf{1)} We have designed a two-stage training strategy that combines knowledge distillation and fine-tuning to train a parameter-efficient VPR model based on visual foundation models. Additionally, we propose the Distillation Recovery Module (DRM) to minimize knowledge loss during the knowledge distillation process to the greatest extent possible. \textbf{2)} We design a Top-Down-attention-based Deformable Aggregator (TDDA), which controls the deformation of local ROIs through global semantic information and demonstrates better adaptability to irregular structures and regions compared with existing aggregators centered on spatial pooling. \textbf{3)} Extensive experiments show that our $D^{2}$-VPR can deliver competitive performance compared with SOTA methods on popular benchmarks while achieving significant reductions in parameter count and storage consumption.

\section{Related Works}
VPR methods fall into two main categories. One‑stage VPR generates a single global descriptor per image by aggregating local features—early methods used handcrafted features such as SURF~\cite{SURF}, while more recent approaches employ deep learning architectures like NetVLAD~\cite{netvlad}, MixVPR~\cite{mixvpr}, and AnyLoc~\cite{keetha2023anyloc}. These methods facilitate efficient, end-to-end retrieval. Two‑stage VPR first retrieves candidates using global descriptors and then refines the top-\(K\) results through local feature matching. Representative examples include Patch-NetVLAD~\cite{patchvlad}, which improves precision at the cost of additional computation. Since our goal is to achieve a better trade-off between parameter-efficiency and performance, our method focuses on one-stage VPR.

\noindent\textbf{VPR Based on Visual Foundation Models}. With the emergence of visual foundation models trained on massive amounts of data via unsupervised learning, a new wave of VPR methods has adopted DINOv2~\cite{oquab2023dinov2} as the backbone to build more robust descriptors. For instance, DINO‑Mix~\cite{huang2024dino} combines DINOv2 with a multi-layer-perceptron-based aggregation, significantly improving performance under challenging illumination and seasonal variations. SALAD~\cite{salad} fine-tunes DINOv2 and leverages optimal-transport-based aggregation to set new SOTA results for one-stage VPR.  EffoVPR~\cite{tzachor2024effovpr} extracts internal attention features from frozen DINOv2 layers to create compact yet discriminative descriptors. EDTformer~\cite{EDTformer} uses a lightweight decoder transformer with low-rank adaptation to refine DINOv2 features efficiently, while SciceVPR~\cite{wan2025scicevpr} enhances cross-image correlation and multilayer fusion to boost retrieval robustness. Although these VPR techniques, backed by powerful foundation models, mitigate challenges such as appearance changes, viewpoint variations, and perceptual aliasing to a certain extent, their large parameter size and heavy computational demand greatly limit deployment on edge or resource-constrained devices.

\begin{figure}[t]
    \centering
    \includegraphics[width=1\linewidth]{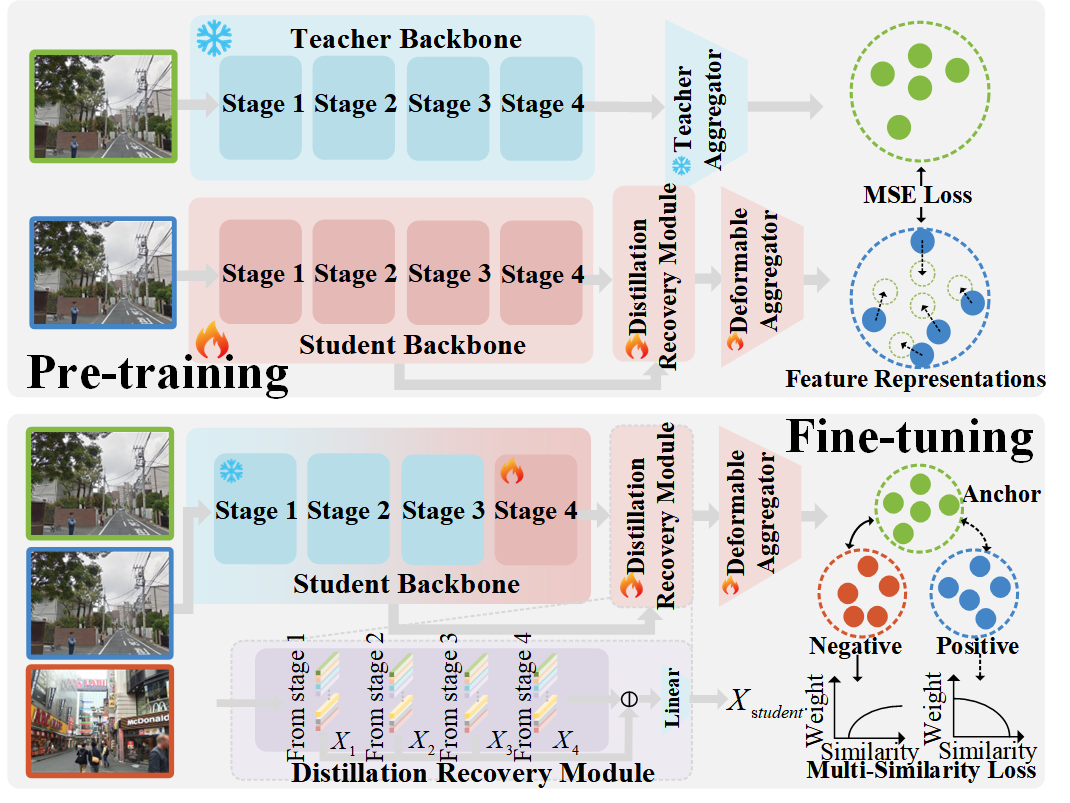}
    \caption{ Two training stages of our VPR model.}
\label{pipeline}
\vspace{-0.3cm}
\end{figure}

\noindent\textbf{Parameter-efficient VPR} aims to balance model compactness and retrieval performance. LPS‑VPR~\cite{nie2024efficient} uses a pooling-centric saliency encoder to fuse multi-scale CNN features, while EPSA‑VPR~\cite{nie2025epsa} introduces patch saliency–weighted aggregation on ResNet-50 for compact, robust descriptors. Though lightweight, these methods lag in performance as their backbone features lack the strong generalization of visual foundation models. TeTRA-VPR~\cite{grainge2025tetravpr}, with similar goals to ours, also uses knowledge distillation for efficient foundation-model-based VPR. However, the key difference lies in that their core approach is to binarize the model parameters, which still impairs the performance of visual foundation models, leading to a slight performance drop between the binarized model and the original one.

Our approach aims to bridge the gap between existing efficient VPR models and those based on large-scale pretrained visual foundation models by integrating their respective strengths. It maintains a lightweight architecture while effectively harnessing the powerful representations learned by foundation models, thereby achieving a more favorable balance between parameter-count and retrieval performance.

\section{Methodology}
\subsection{Two-stage Training Strategy}

As shown in Figure~\ref{pipeline}, the training process of our model is divided into two stages: a pre-training stage centered on knowledge distillation (Mean Squared Error (MSE) loss is used here), and a fine-tuning stage centered on the multi-similarity loss~\cite{wang2019multi}. 

\noindent\textbf{Knowledge Distillation Based Pre-training} aims to compress and transfer the knowledge of a teacher model based on a visual foundation model to a student model, thereby achieving effective parameter initialization, particularly for the backbone. This process enables the lightweight student model to approximate the teacher model’s semantic understanding and representation capacity while maintaining a lower computational burden. Specifically, we employ CricaVPR \cite{crica} with a DINOv2-base backbone as the teacher model, and select DINOv2-small as the backbone of our $D^2$-VPR (student model). To minimize knowledge loss during knowledge transfer, a distillation recovery module is introduced to align the feature dimensions of both models. At this stage, we optimize the entire parameter set of the student model by minimizing the MSE loss between the output features of the teacher and student models, ensuring the effective transfer of representational knowledge.

\noindent\textbf{Fine-tuning} stage is to further enhance the representational capacity of the student model by adapting it to the VPR task. This is achieved by updating a limited number of parameters, primarily from the deeper layers closer to the output, while keeping most of the parameters of the backbone fixed. Specifically, we adopt the multi-similarity loss \cite{wang2019multi} as the optimization loss as shown in Equation \ref{ms}, which is widely used in metric learning for VPR tasks. During this stage, the parameters of the first three stages of the backbone are frozen to prevent the catastrophic forgetting of the knowledge transferred during pre-training, and only the last stage of the backbone and the following layers are updated.

\begin{equation}
\begin{split}
\mathcal{L}_{\text{MS}} = \frac{1}{N} \sum_{i=1}^N \bigg[ 
& \frac{1}{\alpha} \log\left(1 + \sum_{j \in P_i} \exp\left(-\alpha (s_{ij} - \lambda)\right)\right) \\
+ & \frac{1}{\beta} \log\left(1 + \sum_{k \in N_i} \exp\left(\beta (s_{ik} - \lambda)\right)\right)
\bigg]
\end{split}
\label{ms}
\end{equation}
where \( N \) is the number of training samples, \( s_{ij} = \langle F_i, F_j \rangle \) denotes the cosine similarity between features, \( P_i \) and \( N_i \) represent the mined positive and negative sets for anchor \( i \), \( \alpha \) and \( \beta \) are weighting hyperparameters, and \( \lambda \) is a margin.

\subsection{Model Architecture}

As shown in Figure~\ref{pipeline}, the overall model architecture is divided into three parts. 1) The backbone, a small visual foundation model, is responsible for extracting features from images. 2) The distillation recovery module fuses the features of the backbone from shallow to deep layers and aligns them with the feature dimensions of the teacher model, so as to avoid losses in knowledge transfer to the greatest extent. 3) The top-down-attention-based deformable aggregator, on the other hand, further compresses the extracted features to form a compact feature representation related to the VPR task.

\noindent\textbf{Backbone}. We adopt a vision transformer backbone (DINOv2-small) to extract features from the input image. The output $X_{4}\in \mathbb{R}^{B \times (1+P) \times D/2}$ from the final transformer layer (stage 4) consists of a global class token and patch-level tokens:

\begin{equation}
X_{4} = \{ f_{\text{cls}}, f_1, f_2, \dots, f_{P} \} 
\end{equation}
where \( f_{\text{cls}} \in \mathbb{R}^{D/2} \) is the class token, \( f_i \in \mathbb{R}^{D/2} \) are the spatial patch tokens, and $D/2$ is the feature dimension ($D$ for tokens' dimension of teacher backbone). 

\noindent\textbf{Distillation Recovery Module}. To align the feature output dimensions of the teacher model and the student model, thus reducing knowledge loss during pre-training, we introduce a distillation recovery module at the end of the student backbone as shown in Figure~\ref{pipeline}. This module recovers feature dimensions consistent with those of the teacher model by fusing features from shallow to deep layers. Specifically, since the teacher model uses DINOv2-base with an output feature dimension ($D$) that is twice that of our student model ($D/2$), we bridge this gap by concatenating hierarchical features from all four stages of the student backbone in a shallow-to-deep order and projecting them into the teacher's feature space via a linear transformation:

\begin{equation}
X_{\text{student}} = \text{Linear} \left( \text{Concat}(X_{1}, X_{2}, X_{3}, X_{4}) \right)
\end{equation}
where \( X_{i} \) denotes the output token sequence from the \( i \)-th stage of the student backbone. This alignment mechanism constructs a shared knowledge distillation space and enables the student model to mimic the teacher's semantic representations while maintaining low computational complexity. Finally, \( X_{\text{student}} \in \mathbb{R}^{B \times (1+P) \times D} \) is rearranged into a spatial feature map \( F \in \mathbb{R}^{B \times D \times H \times W} \) and class token as \( f_{\text{class}} \in \mathbb{R}^{B \times D} \) for subsequent aggregation.

\noindent\textbf{Top-down-attention-based Deformable Aggregator}

\begin{figure}[t]
    \centering
    \includegraphics[width=1\linewidth]{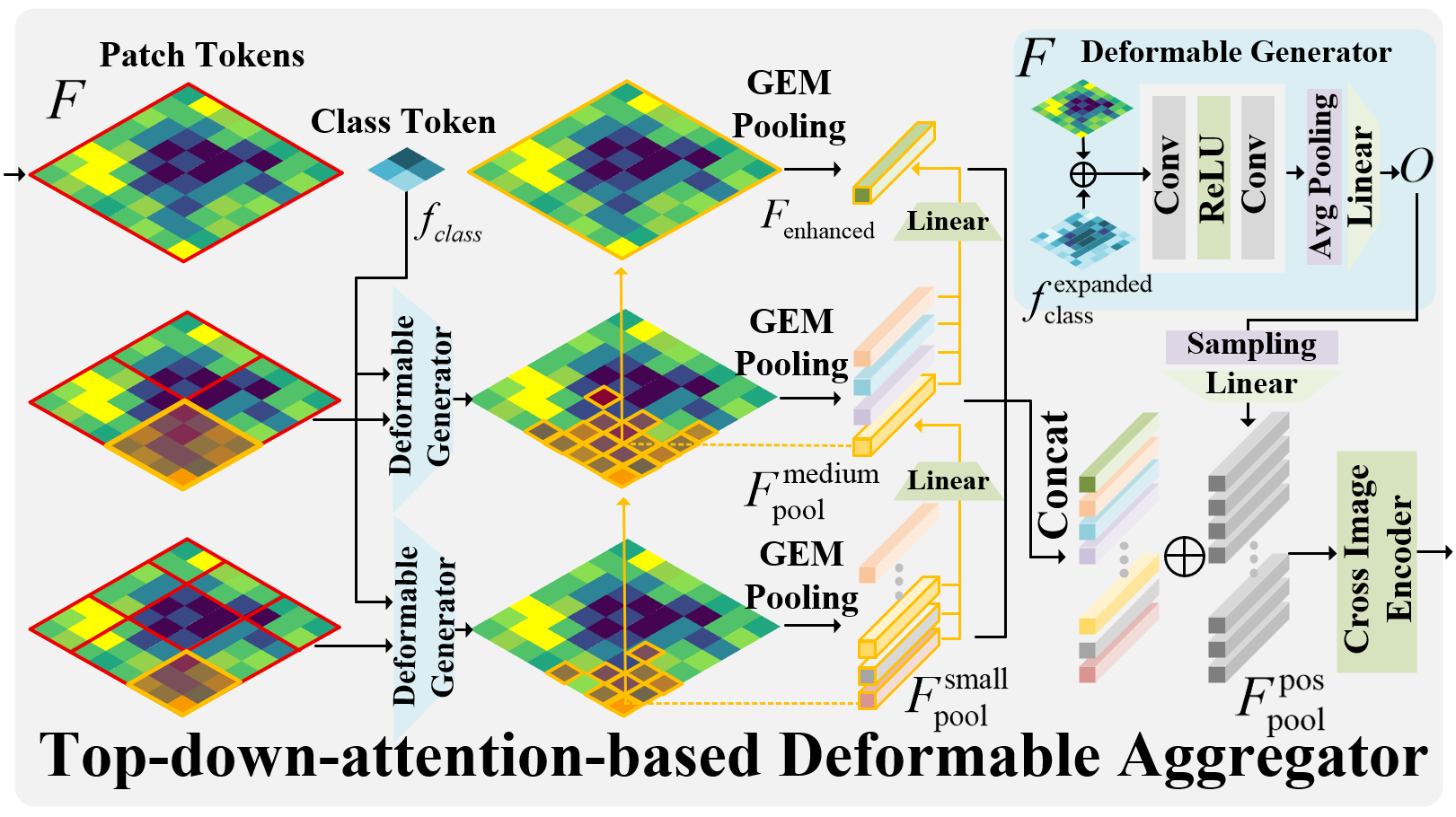}
    \caption{Top-down-attention-based deformable aggregator.}
\label{TDDA}
\vspace{-0.3cm}
\end{figure}

Top-down neural attention \cite{GILBERT2007677} plays a crucial role in human visual perception. It suggests that the brain first constructs a rapid and abstract interpretation of a scene, which is then employed to guide and refine the processing of incoming sensory signals. This ultimately results in more accurate recognition of the positions, shapes, and categories of objects. As shown in Figure~\ref{TDDA}, the design of our TDDA follows this concept by utilizing global and semantic information to drive corresponding deformations of ROIs, enabling the model to better focus on irregular geometric regions, and is divided into the following components: multi-scale pyramid ROI, top-down deformable region pooling, hierarchical down-top fusion, and deformation-aware position embedding.

\noindent\textbf{\textit{Multi-Scale Pyramid ROI.}}  
To align with homogeneous knowledge distillation (the teacher and student models have the same architecture) and thereby improve distillation efficiency \cite{gou2021knowledge}, we adopt the same multi-scale pooling strategy as employed in the teacher's aggregator (CricaVPR). Specifically, we divide the input image into multiple spatial regions at different granularities, including one global region, four medium-scale regions, and nine fine-grained small regions as indicated by the red line division in Figure \ref{TDDA}. Each region is defined by an ROI \( (x_1, y_1, x_2, y_2) \) over the backbone feature map and is processed through the subsequent deformable region pooling mechanism.

\noindent\textbf{\textit{Top-down Deformable Region Pooling}}. To improve the model's perception flexibility toward irregular and non-rigid spatial regions, we propose a top-down-attention-based deformable region pooling method. This mechanism allows each region to adaptively adjust its receptive field based on the fusion of local content (ROI patch token) and global semantic context (class token), improving robustness to viewpoint, scale, and structural variations. Specifically, for an ROI, we construct a sampling grid \( \mathcal{G}_{\text{base}} \in \mathbb{R}^{H' \times W' \times 2} \) over this region, where $H'$ and $W'$ represent the resolution of the sampling grid. To incorporate global context, the class token \( f_{\text{class}} \) is spatially broadcast and concatenated with the full feature map along the channel dimension:

\begin{equation}
F_{\text{fused}} = \text{Concat}(F, f_{\text{class}}^{\text{expanded}})
\end{equation}
where \( F_{\text{fused}} \in \mathbb{R}^{B \times 2D \times H \times W} \). Then, a deformable generator (contains two CNN layers as shown in the top-right corner of Figure \ref{TDDA}) is applied to predict four transformation parameters for each spatial location:

\begin{equation}
O = \text{Deformable-Generator}(F_{\text{fused}})
\end{equation}
where $O \in \mathbb{R}^{B \times 4 \times H \times W} $ contains horizontal and vertical offsets \( (\Delta x, \Delta y) \), and raw scaling factors \( (s_w, s_h) \). To obtain localized transformation parameters for each region, we sample the offset and scaling fields at the base grid positions using bilinear interpolation:

\begin{equation}
\tilde{O}_{\text{ROI}} = \text{GridSample}(O, \mathcal{G}_{\text{base}})
\end{equation}
where \( \tilde{O}_{\text{ROI}} \in \mathbb{R}^{B \times H' \times W' \times 4} \) denotes the transformation parameters ( $\tilde{\Delta x}, \tilde{\Delta y}, \tilde{ s_w}, \tilde{s_h}$) at each sampled point of the ROI region. The sampled parameters are subsequently used to deform the base grid coordinates as follows:

\begin{equation}
\begin{aligned}
x_{\text{deform}} &= x_{\text{center}} + (x_{\text{rel}} \cdot \tilde{ s_w} + \tilde{\Delta x}) \cdot \frac{w}{2} \\
y_{\text{deform}} &= y_{\text{center}} + (y_{\text{rel}} \cdot \tilde{s_h} + \tilde{\Delta y}) \cdot \frac{h}{2}
\end{aligned}
\end{equation}
where \( (x_{\text{rel}}, y_{\text{rel}}) \in [-1, 1] \) are the normalized coordinates from the base grid, and \( (x_{\text{center}}, y_{\text{center}}, w, h) \) denote the center and size of the ROI. $x_{\text{deform}}$ and $y_{\text{deform}}$ constitute the deformable sampling grid $\mathcal{G}_{\text{deform}}$, which is then applied to the original feature map using bilinear interpolation:

\begin{equation}
F_{\text{region}} = \text{GridSample}(F, \mathcal{G}_{\text{deform}})
\end{equation}

Each sampled region feature map \( F_{\text{region}} \in \mathbb{R}^{B \times D \times H' \times W'} \) (as shown by the orange irregular region in Figure \ref{TDDA}) is then aggregated into a compact vector representation using Generalized Mean Pooling (GeM)~\cite{radenovic2018fine}, which introduces a learnable parameter \( p \) to balance between average and max pooling:

\begin{equation}
F_{\text{pool}} = \left( \frac{1}{H'W'} \sum_{i=1}^{H'} \sum_{j=1}^{W'} F_{\text{region}}^{(p)}[ :, :, i, j] \right)^{1/p}
\end{equation}

\noindent\textbf{\textit{Hierarchical Down-top Fusion.}} After obtaining the deformed region pooled features driven by the semantic class tokens, we design a down-top fusion strategy that progressively enhances the global feature representation by incorporating fine-grained local deformable information as shown by the orange arrow in Figure \ref{TDDA}, thereby forming a bidirectional interaction mechanism. Specifically, each deformable region—whether at the medium or global level—is refined by aggregating its spatially neighboring lower-level regions based on deformed region centers, followed by a linear projection and residual addition. This unified fusion process can be formulated as:
\begin{equation}
F_{\text{enhanced}} = F^{\text{top}}_{\text{pool}} + \text{Linear} \left( \frac{1}{N} \sum_{i=1}^{N} F^{i}_{\text{pool}}\right)
\end{equation}
where \( F^{\text{top}}_{\text{pool}} \) is the top target region feature to be enhanced (e.g., medium or global), \( \{F^i_{\text{pool}}\}_{i=1}^{N} \) are its associated down (e.g., medium or small) neighboring region pooled features.

\noindent\textbf{\textit{Deformation-Aware Position Embedding.}}  
Deformable region pooling disrupts the spatial regularity of fixed-grid regions, making it difficult for the model to retain explicit awareness of the original location and size of each region (before deformation, the ROI positions and sizes are fixed, and the ordered feature arrangement enables the model to easily perceive this information). To address this, we embed deformation-specific geometric information for each region using its post-deformation center and size. Specifically, for a region, $\tilde{O}_{\text{ROI}}$ is then projected through a linear layer and added to the region feature to inject geometric context:

\begin{equation}
F_{\text{pool}}^{\text{pos}} = F_{\text{pool}} + \text{Linear}(\tilde{O}_{\text{ROI}})
\end{equation}

After injecting deformation-aware positional information into each region feature, all region descriptors from multiple scales are concatenated to form a unified feature set. Following \cite{crica}, we feed this sequence into a cross-image transformer encoder. The output sequence is then flattened and subjected to L2 normalization to obtain the final descriptor, yielding compact representations.

\begin{table}[t]
\setlength{\extrarowheight}{0pt}
\setlength{\aboverulesep}{0pt}
\setlength{\belowrulesep}{0pt}
\small
\centering
\setlength{\tabcolsep}{1.4pt}
\renewcommand{\arraystretch}{1}
\begin{tabular}{@{}l||c||c||c@{}}
\toprule[1.5px]
\multirow{2}{*}{Dataset} & \multirow{2}{*}{Description} & \multicolumn{2}{c}{Number} \\
\cmidrule{3-4}
 & & Database & Queries \\
\midrule
Pitts30k & urban, panorama & 10,000 & 6,816  \\
Pitts250k & urban, panorama & 83952 &8280  \\
MSLS-val & urban, suburban & 18,871 & 740  \\

Nordland & natural, seasonal & 27,592 & 27,592 \\

AmsterTime & very long-term & 1,231 & 1,231 \\
SPED& various scenes & 607 & 607  \\
\bottomrule[1.5px]
\end{tabular}
\caption{Summary of the test datasets in experiments.}
\label{tab:datasets}
\end{table}

\begin{table*}[t]
\setlength{\extrarowheight}{0pt}
\setlength{\aboverulesep}{0pt}
\setlength{\belowrulesep}{0pt}
\small
\centering
\setlength{\tabcolsep}{1.4pt}
\renewcommand{\arraystretch}{1.2}
{

\begin{tabular}{@{}l||ccc||ccc||ccc||ccc||ccc||ccc}
\toprule[1.5px]
\multirow{2}{*}{Method} 

& \multicolumn{3}{c||}{Pitts30k} 
& \multicolumn{3}{c||}{MSLS-val} 
& \multicolumn{3}{c||}{SPED}
& \multicolumn{3}{c||}{Pitts250k}
& \multicolumn{3}{c||}{AmsterTime}
& \multicolumn{3}{c}{Nordland} \\
\cline{2-19}
& R@1 & R@5 & R@10
& R@1 & R@5 & R@10
& R@1 & R@5 & R@10
& R@1 & R@5 & R@10
& R@1 & R@5 & R@10
& R@1 & R@5 & R@10 \\
\hline

Cosplace  

& 90.9 & 95.7 & 96.7
& 87.4 & 94.1 & 94.9
& - & - & -
& 92.3 & 97.4 & 98.4
& 47.7 & \underline{69.8} & 75.8
& 71.9 & 83.8 & 88.1
\\

MixVPR   

& 91.5 & 95.5 & 96.3
& 87.2 & 93.1 & 94.3
& - & - & -
& \underline{94.1} & \underline{98.2} & \textbf{98.9}
& 40.2 & 59.1 & 64.6
& 76.2 & 86.9 & 90.3
\\

EigenPlaces     

& \textbf{92.5} & \textbf{96.8} & \textbf{97.6}
& 89.1 & 93.8 & 95.0
& - & - & -
& \underline{94.1} & 97.9 & \underline{98.7}
& \underline{48.9} &69.5 & \underline{76.0}
& 71.2 & 83.8 & 88.1
\\

LPS-VPR    

& 91.8 & \underline{96.0} & \underline{96.8}
& \underline{89.9} & \underline{94.2} & 95.0
& \underline{84.8} & \textbf{93.9} & \underline{95.7}
& \underline{94.1} & 98.0 & \textbf{98.9}
& - & - & -
& - & - & -
\\

$\text{BoQ}^\dagger$    

& \underline{92.0} & 95.6 & 96.6
& 86.6 & 92.3 & 93.5
& 82.7 & \underline{91.3} & 94.2
& \textbf{94.4} & 97.9 & \underline{98.7}
& 42.0 & 60.5 & 66.5
& \textbf{78.9} & \underline{88.5} & \underline{91.5}
\\

EPSA-VPR

& - & - & -
& 89.3 & 93.8 & \underline{95.4}
& 84.5 & 93.7 & \textbf{96.0}
& 93.9 & 97.9 & \underline{98.7}
& - & - & -
& - & - & -
\\

Clus-VPR

& 90.8 & 95.2 & 96.6
& 82.7 & 88.5 & 92.4
& - & - & -
& 92.4 & 96.9 & 97.6
& - & - & -
& - & - & -
\\

$D^{2}$-VPR$_{\text{no encoder}}$           

& 91.7 & 95.8 & \underline{96.8}
& \textbf{90.7} & \textbf{95.4} & \textbf{96.4}
& \textbf{86.0} & 92.9 & 94.1
& \textbf{94.4} & \textbf{98.3} & \textbf{98.9}
& \textbf{49.1} & \textbf{70.7} & \textbf{76.1}
& \underline{77.1} & \textbf{88.6} & \textbf{91.9}
\\
\bottomrule[1.5px]
\end{tabular}}
\caption{Comparison with similar-scale SOTA methods on popular benchmarks. The best results are highlighted in \textbf{bold} and the second best are \underline{underlined}. $\text{BoQ}^\dagger$ is re-evaluated using the officially provided weights. The input image size is set to 224×224, and the feature dimension is reduced from 16384 to 4096 using PCA for a fairer comparison. - for not reported. Results of Cosplace and MixVPR are reported from EigenPlaces. Here, $D^{2}$-VPR does not use cross-image encoder.}
\label{tab:compare_efficient}
\end{table*}

\begin{table}[t]
\setlength{\extrarowheight}{0pt}
\setlength{\aboverulesep}{0pt}
\setlength{\belowrulesep}{0pt}
\small
\centering
\setlength{\tabcolsep}{1.4pt}
\renewcommand{\arraystretch}{1}
{

\begin{tabular}{@{}l||c|c|c|c@{}}
\toprule[1.5px]
\multirow{2}{*}{Method} 
& \multirow{2}{*}{\scriptsize Dim.}
& \multirow{2}{*}{\scriptsize Backbone}
& \scriptsize Image 
& \scriptsize Param.  \\

& & &\scriptsize Size &(M) \\
\hline

Cosplace~$_{{\text{CVPR' 2022}}} $     
& 2048 & ResNet50 & No Resize & 27.70 \\

MixVPR~$_{{\text{WACV' 2023}}} $     
& 4096 & ResNet50 & No Resize & \textbf{10.88} \\

EigenPlaces~$_{{\text{ICCV' 2023}}} $     
& 2048 & ResNet50 & No Resize & 27.70 \\

LPS-VPR~$_{{\text{RAL' 2024}}} $     
& 2048 & ResNet50 & 640×480 & 29.71\\

$\text{BoQ}$~$_{{\text{CVPR' 2024}}} $     
& 4096& ResNet50 & 224×224 &  \underline{23.84}\\

EPSA-VPR~$_{{\text{JVCI' 2025}}}$    
& 1024 & ResNet50 & - & 27.71 \\

Clus-VPR~$_{{\text{TAI' 2025}}}$    
& 4096& CWTNet &  640×480 & 53.12\\

$D^{2}$-VPR$_{\text{no encoder}}$           
& 4096 & Dinov2 & 224×224 & 27.21 \\

\bottomrule[1.5px]
\end{tabular}}
\caption{Detailed information of the similar-scale SOTA methods in the comparison. Clus-VPR and EPSA-VPR report only the parameters of their aggregators. Therefore, we use their full ResNet50 version and report 23.51M parameters for the backbone. Note that MixVPR and BoQ utilize a cropped ResNet-50 as their backbone, with the parameter count of the backbone network being less than 23.51M.}
\label{tab:compare_efficient2}
\end{table}

\section{Experiments}
\subsection{Benchmarks}
We conduct experiments on multiple VPR benchmark datasets that exhibit viewpoint variations, environmental condition changes, and perceptual aliasing challenges. Table~\ref{tab:datasets} summarizes these datasets: Pitts30k and Pitts250k~\cite{pitts} primarily feature significant viewpoint changes; MSLS~\cite{msls} spans urban, suburban and natural scenes captured several years with diverse visual variations; SPED~\cite{SPED} comprises surveillance camera imagery. Additionally, we include challenging datasets: Nordland (seasonal variations)~\cite{nordland} and AmsterTime (long-term changes)~\cite{amstertime}.

We employ Recall@N (R@N) as the evaluation metric, measuring the percentage of queries where at least one top-N retrieved database image is within a ground truth threshold. Following standard protocols~\cite{msls,pitts}, thresholds are: 25m + 40$^{\circ}$ for MSLS; 25m for Pitts30k, Pitts250k and SPED; $\pm$10 frames for Nordland; and unique counterpart matching for AmsterTime.

\begin{table*}[t]
\setlength{\extrarowheight}{0pt}
\setlength{\aboverulesep}{0pt}
\setlength{\belowrulesep}{0pt}
\small
\centering
\setlength{\tabcolsep}{1.4pt}
\renewcommand{\arraystretch}{1.2}
{
\begin{tabular}{@{}l||ccc||ccc||ccc||ccc||ccc||ccc}
\toprule[1.5px]
\multirow{2}{*}{Method} 

& \multicolumn{3}{c||}{Pitts30k} 
& \multicolumn{3}{c||}{MSLS-val} 
& \multicolumn{3}{c||}{SPED}
& \multicolumn{3}{c||}{Pitts250k}
& \multicolumn{3}{c||}{AmsterTime}
& \multicolumn{3}{c}{Nordland} \\
\cline{2-19}
& R@1 & R@5 & R@10
& R@1 & R@5 & R@10
& R@1 & R@5 & R@10
& R@1 & R@5 & R@10
& R@1 & R@5 & R@10
& R@1 & R@5 & R@10 \\
\hline

$\text{CricaVPR}^{*}$
& \textbf{94.9} & \textbf{97.3} & \textbf{98.2}
& 90.0 & 95.4 & 96.4
& \textbf{91.9} & 95.7 & 96.7
& \underline{97.5} & \textbf{99.4} & \textbf{99.7}
& \textbf{64.7} & \textbf{82.8} & \textbf{87.5}
& \textbf{90.7} & \textbf{96.3} & \textbf{97.6} \\

SALAD
& 91.6 & 95.7 & 97.1
& 90.5 & 95.5 & 96.2
& 90.8 & 95.2 & \underline{96.9}
& 94.6 & 98.1 & 98.9
& 53.4 & 75.0 & 79.9
& 81.2 & 91.1 & 94.0 \\

BoQ
& 93.1 & 96.5 & 97.5
& \textbf{91.6} & 95.9 & 96.8
& \underline{91.8} & 95.6 & 96.5
& 95.9 & 98.8 & 99.4
& 57.6 & 76.9 & 81.9
& 85.8 & 93.6 & 95.9 \\

$\text{SuperVLAD}^{*}$
& \underline{94.1} & \textbf{97.3} & \underline{98.0}
& 90.7 & 95.9 & 96.8
& 90.9 & 95.6 & 96.5
& 96.1 & \underline{99.0} & \underline{99.5}
& 60.0 & 80.3 & 84.4
& \underline{88.6} & \underline{94.7} & \underline{96.5} \\

Sela-global
& 90.2 & 96.1 & 97.1
& 87.7 & 95.8 & 96.6
& 84.5 & 91.8 & 93.9
& 92.8 & 98.0 & 98.9
& 41.5 & 62.1 & 69.3
& 72.3 & 89.4 & 94.4 \\

FoL-global
& 92.6 & \underline{96.9} & 97.7
& 90.4 & 95.7 & \underline{96.9}
& 90.6 & \textbf{96.2} & \underline{96.9}
& 95.3 & 98.8 & 99.4
& 54.1 & 76.0 & 81.0
& 74.7 & 86.9 & 91.0 \\

EDTformer
& 92.9 & 96.8 & 97.8
& \underline{91.5} & \textbf{96.4} & 96.6
& 90.9 & 95.4 & 96.7
& 95.5 & 98.7 & 99.3
& 58.2 & \underline{80.8} & 84.8
& 81.0 & 91.2 & 94.1 \\

$D^{2}$-VPR$_{\text{no encoder}}$           

& 91.7 & 95.8 & 96.8
& 90.7 & 95.4 & 96.4
& 86.0 & 92.9 & 94.1
& 94.4 & 98.3 & 98.9
& 49.1 & 70.7 & 76.1
& 77.1 & 88.6 & 91.9\\

$D^{2}\text{-VPR}^{*}$$_{\text{encoder}}$           
& \textbf{94.9} & \textbf{97.3} & \underline{98.0}
& \textbf{91.6} & \underline{96.1} & \textbf{97.2}
& 90.9 & \underline{96.0} & \textbf{97.4}
& \textbf{97.8} & \textbf{99.4} & \textbf{99.7}
& \underline{62.9} & \underline{80.8} & \underline{85.3}
& 86.6 & 94.1 & 96.0 \\
\bottomrule[1.5px]
\end{tabular}}
\caption{We compare our method with larger-scale SOTA methods on popular benchmarks. To ensure a fair comparison, the evaluation resolution is set to 224×224. Except for CricaVPR, all other methods are originally evaluated at higher resolutions, so we re-evaluate them using their source code and provided weights. Therefore, the results may differ from those reported in the original papers. Methods with the suffix ‘-global’ correspond to the first stage of the two-stage approach.* for using cross-image encoder. }
\label{tab:compare_SOTA}
\end{table*}

\begin{table}[t]
\setlength{\extrarowheight}{0pt}
\setlength{\aboverulesep}{0pt}
\setlength{\belowrulesep}{0pt}
\small
\centering
\setlength{\tabcolsep}{1.4pt}
\renewcommand{\arraystretch}{1.2}
{

\begin{tabular}{@{}l||c|c|c|c}
\toprule[1.5px]
\multirow{2}{*}{Method} 
& \multirow{2}{*}{\scriptsize Dim.}
& \multirow{2}{*}{\scriptsize Dinov2}
& \scriptsize Image 
& \scriptsize Param.   \\

& & & Size & (M)  \\
\hline
CricaVPR~$_{{\text{CVPR' 2024}}}$ 
&4096  &Base  &224×224& 106.76   \\

SALAD~$_{{\text{CVPR' 2024}}}$ 
&  8448&Base  &224×224 &87.99   \\

BoQ~$_{{\text{CVPR' 2024}}}$ 
&  12288& Base &224×224 & 95.21   \\

SuperVLAD~$_{{\text{NIPS' 2024}}}$ 
& 3072 &Base   &224×224& 97.61   \\

Sela-global~$_{{\text{ICLR' 2024}}}$ 
&  1024& Large  &224×224& 357.43    \\

FoL-global~$_{{\text{AAAI' 2025}}}$ 
&  8448& Large &224×224 & 308.83    \\

EDTformer~$_{{\text{TCSVT' 2025}}}$ 
& 4096 &Base  &224×224 & 96.96   \\

$D^{2}$-VPR~$_{{\text{no encoder}}}$        
&  4096& Small  &224×224& \textbf{27.21}  \\

$D^{2}$-VPR~$_{{\text{encoder}}}$        
&  4096& Small  &224×224& \underline{38.24}  \\
\bottomrule[1.5px]
\end{tabular}}
\caption{Detailed information of the larger-scale SOTA methods in the comparison. }
\label{tab:compare_SOTA2}
\end{table}

\subsection{Implementation Details} 

Our training follows a two-stage strategy. In the knowledge distillation stage, we use CricaVPR with DINOv2-base as the teacher model, and our $D^{2}$-VPR (student model) uses DINOv2-small (initialized with pretrained weights) as the backbone. During this stage, all parameters of the student model are trained. We use the ADAM~\cite{kingma2014adam} optimizer with a batch size of 8 and a learning rate of 2.5e-5. In the fine-tuning stage, we freeze the first 3/4 layers of the backbone and train the remaining parameters. This stage uses the ADAMW~\cite{loshchilov2017fixing} optimizer with a batch size of 128 and a learning rate of 2e-4. Both stages are trained on the GSV-Cities dataset~\cite{ali2022gsv}, a large-scale urban location dataset collected via Google Street View. Each batch consists of 4 images, and the input image size for both training and evaluation is set to 224×224. All experiments are conducted on an RTX 3090 GPU, with PyTorch 2.3.0 and Python 3.10. Our model outputs 10752-dimensional global features, and following~\cite{crica}, we apply principal component analysis (PCA) for dimensionality reduction to 4096 dimensions.

\subsection{Comparisons with State-of-the-art Methods}

\begin{figure}[t]
    \centering
    \includegraphics[width=1\linewidth]{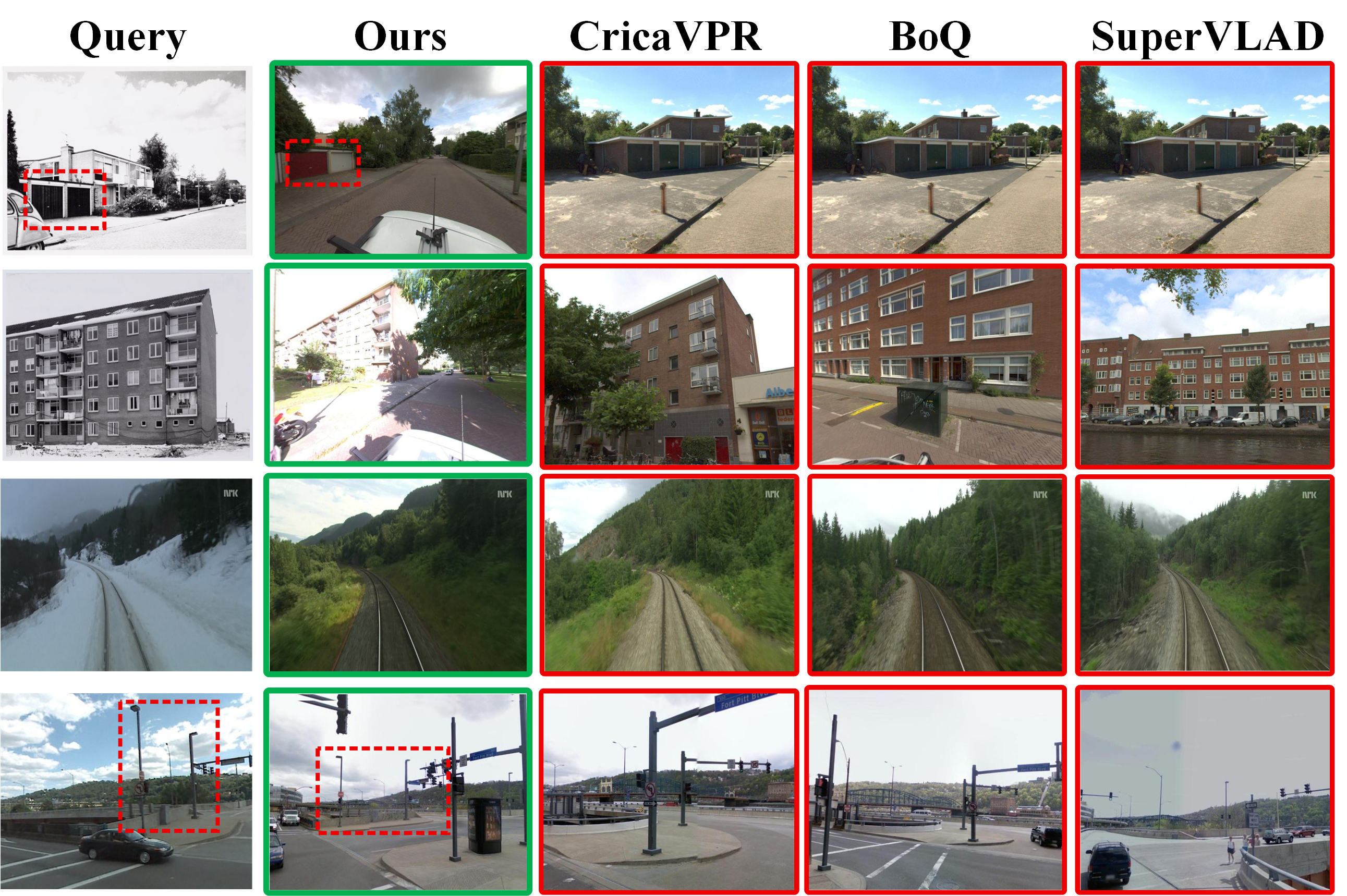}
    \caption{ Qualitative VPR comparison results. Our method demonstrates competitive performance compared to these DINOv2-based SOTA models under these challenging cases: long-term appearance changes (first row), drastic lighting variations (second row), perceptual aliasing (third row), and viewpoint changes (fourth row). Green indicates the right match while red is for the wrong one. Key matching regions are highlighted with red dashed boxes.}
\label{Qualitative}
\vspace{-0.3cm}
\end{figure}

\noindent\textbf{\textit{Comparison with Similar-scale SOTA Baselines}}. We conduct evaluations against several similar-scale approaches, including LPS-VPR~\cite{nie2024efficient}, Clus-VPR~\cite{xu2024clusvpr}, EPSA-VPR~\cite{nie2025epsa}, BoQ~\cite{ali2024boq}, Cosplace~\cite{Berton_2022_CVPR}, MixVPR~\cite{Ali-bey_2023_WACV} and EigenPlaces~\cite{Berton_2023_ICCV}. These methods adopt relatively lightweight CNN architectures. As shown in Tables~\ref{tab:compare_efficient} and \ref{tab:compare_efficient2}, our proposed method achieves the best performance on most datasets, particularly on MSLS-val (exceeding the second-best by 0.9 in R@5), AmsterTime (exceeding the second-best by 1.2 in R@5), and Pitts250k, while maintaining a competitive parameter size and using smaller input image resolution of 224×224.

\noindent\textbf{\textit{Comparison with Larger-scale SOTA Baselines}}. We also comprehensively compare our proposed method with SOTA one-stage VPR methods in Tables \ref{tab:compare_SOTA} and \ref{tab:compare_SOTA2}, including CricaVPR~\cite{crica}, SALAD~\cite{salad}, BoQ~\cite{ali2024boq}, SuperVLAD~\cite{lu2024supervlad}, SelaVPR~\cite{sela}, FoL~\cite{wang2025focus}, and EDTformer~\cite{EDTformer}. Note that we have unified the evaluation image resolution to 224×224 rather than a higher resolution, which aligns with our goal of deploying the model on resource-constrained devices. When using the cross-image encoder, the results demonstrate competitive performance, particularly on Pitts30k, MSLS-val, Pitts250k, and AmsterTime. Figure \ref{Qualitative} provides visualizations comparing the retrieval performance of our method with other baselines. However, due to the inherent limitations of the cross-image encoder, we also report comparisons without it. In this setting, our method achieves performance comparable to SALAD on Pitts30k, MSLS-val, and Pitts250k, and performs better than FoL-global on Nordland, indicating that our approach still maintains a reasonable level of competitiveness, despite requiring substantially fewer parameters than the competing methods. 

\begin{figure}
    \centering
    \includegraphics[width=1\linewidth]{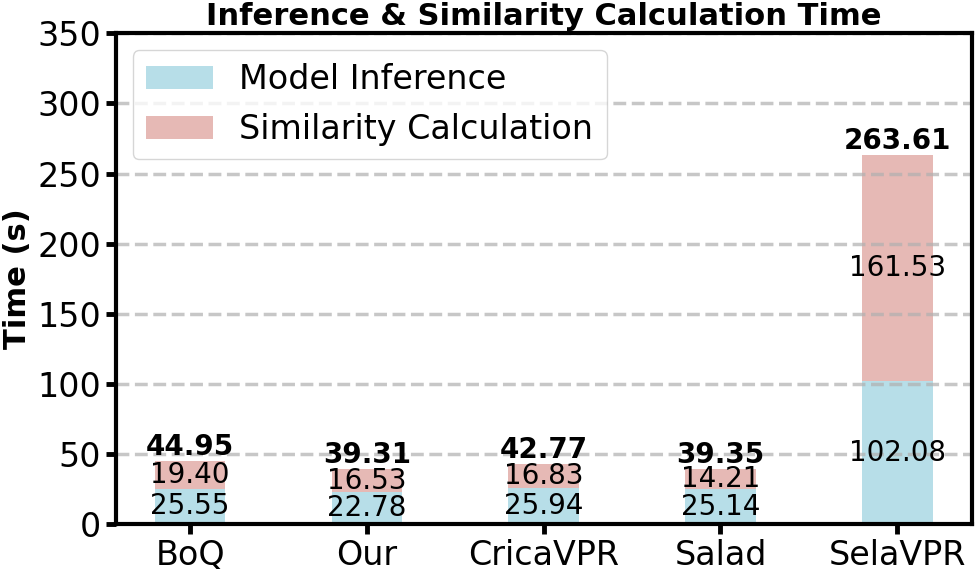}
    \caption{Method comparison of inference and computational speed on AmsterTime. The inference time includes the inference of both the database and the query set. The batch size is 32. SelaVPR performs calculations in a two-stage manner. PCA is not used here.}
\label{inf}
\vspace{-0.3cm}
\end{figure}

\begin{table}
\centering
\setlength{\tabcolsep}{3mm}
\renewcommand\arraystretch{1.2}

\begin{tabular}{l||ccc} 
\toprule[1.5px]
\multirow{2}{*}{Configurations} & \multicolumn{3}{c}{MSLS-val}  \\
\cline{2-4}
                                & R@1 & R@5 & R@10                    \\ 
\hline

Baseline                        &  85.1&  93.5&95.3                                             \\ 

$+ \text{distillation}$           & 88.4 & 95.0 & 96.4                                          \\

$+ \text{finetune}$           & 91.6 & 96.1 & 97.2                                             \\

\hline

Baseline                        & 89.9&  95.0&95.7                                             \\

$+ \text{DRM}$            & 90.5 & 95.5 & 96.2                                           \\

$+  \text{TDDA}$            & 91.6 & 96.1 & 97.2                         \\

\bottomrule[1.5px]
\end{tabular}
\caption{Ablation study on the MSLS-val benchmark.}
\label{ablation}
\vspace{-2mm}
\end{table}

\noindent\textbf{\textit{Comparison of Inference and Computational Speed}}. As shown in Figure~\ref{inf}, our method achieves competitive inference speed and overall processing time. In terms of similarity calculation time, our approach ranks as the second fastest (here our method does not use PCA and retains the original feature dimension of 10752)—behind SALAD~\cite{salad}, whose descriptor dimensionality is smaller (8448)—and remains comparable to CricaVPR~\cite{crica}.

We observe that the improvement in inference time is not as substantial as expected. This is primarily because the deformable aggregator introduces multiple sampling operations that involve grid generation and bilinear interpolation. These operations cannot be effectively parallelized or fused like standard convolution, thereby constraining the inference speed. We plan to explore more efficient and hardware-friendly designs for this component in future work to improve the overall acceleration.

\subsection{Ablation Study}
\label{ablationsec}

We conduct ablation studies on both training strategies and module designs, as shown in Table \ref{ablation}.

\begin{figure}[t]
    \centering
    \includegraphics[width=1\linewidth]{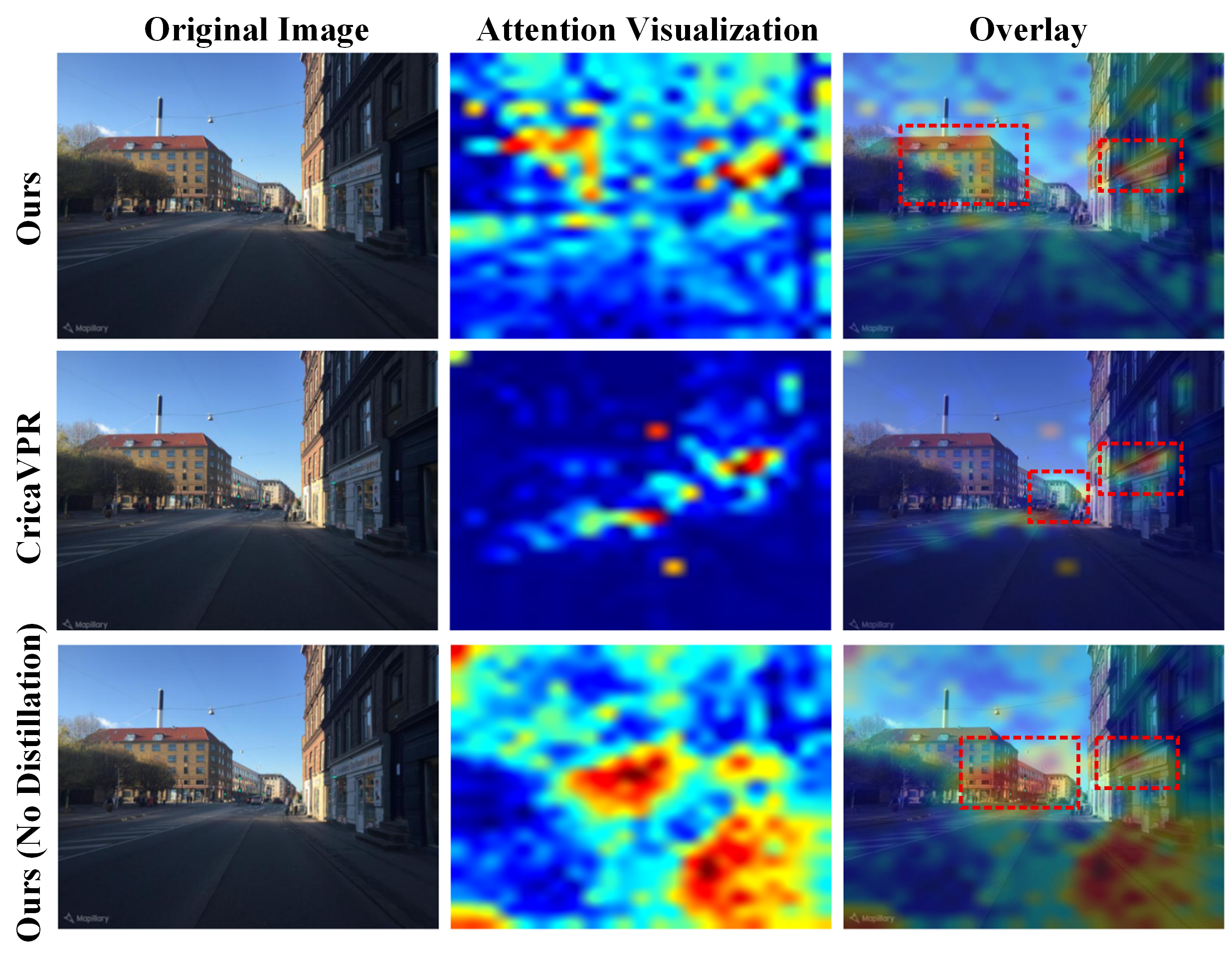}
    \caption{ Visualizations of attention maps from the knowledge distillation ablation study are presented below, with rows corresponding to the student model (Ours), teacher model (CricaVPR), and a baseline without distillation. The non-distilled baseline exhibits the strongest focus on irrelevant regions (e.g., ground and sky), followed by the student model, while the teacher model best suppresses such spurious attention.}
\label{ablationfigure}
\vspace{-0.3cm}
\end{figure}

\noindent\textbf{\textit{Training Strategy}}. Using the baseline (direct fine-tuning without distillation) as reference, introducing knowledge distillation significantly improves performance on MSLS-val (+3.3\%, +1.5\%, +1.1\% on R@1/5/10), indicating that the teacher model’s knowledge is effectively transferred. We also provide distillation comparison visualizations in Figure \ref{ablationfigure}. Further applying fine-tuning yields additional gains (+3.2\%, +1.1\%, +0.8\%), confirming its importance for adapting to the final retrieval task.

\noindent\textbf{\textit{Module Design}}. Adding the distillation recovery module leads to consistent improvement (+0.6\%, +0.5\%, +0.5\%), showing that it effectively fuses shallow and deep backbone features and reduces knowledge-transfer loss. Introducing the deformable aggregator yields the best performance (+1.1\%, +0.6\%, +1.0\%), demonstrating its ability to flexibly handle irregular ROIs and enhance overall retrieval accuracy.

\subsection{Conclusion}

In this work, we present $D^{2}$-VPR, a lightweight visual-foundation-model-based framework that combines knowledge distillation and deformable aggregation to retain the strong representation capabilities of visual foundation models while significantly reducing computational cost. Through a two-stage training strategy and the proposed distillation recovery module, our method effectively bridges the feature gap between teacher and student models. The top-down-attention-based deformable aggregator further enhances adaptability by dynamically adjusting aggregation regions based on global semantics. Extensive experiments demonstrate that $D^{2}$-VPR achieves a favorable balance between performance and efficiency.

\section{Acknowledgments}
This work was supported by the National Key R\&D Program of China under Grant 2024YFF0907404.

\section{Supplementary Material}

This supplementary material provides the following additional content about experimental results and analysis:

A. Influence of Cross Image Encoder

B. Influence of PCA

C. Detials of $D^{2}$-VPR

D. Attention Visualization

E. Details of the Datasets

\subsection{A. Influence of Cross Image Encoder}

\begin{table}[h]
\centering
\setlength{\tabcolsep}{3mm}
\renewcommand\arraystretch{1.2}

\begin{tabular}{l||ccc} 
\toprule[1.5px]
\multirow{2}{*}{Batchsize } & \multicolumn{3}{c}{MSLS-val}  \\
\cline{2-4}
                               & R@1 & R@5 & R@10                    \\ 
\hline

1                        &  90.9&  95.8&96.6                                             \\ 

16           & 90.5 & 96.5 & 96.9                                        \\
32           & 91.1 & 96.5 & 96.8                                          \\

64           & 90.9 & 96.2 & 96.8                                          \\
128          & 91.6 & 96.1 & 97.2                                          \\
256           & 91.1 & 95.9 & 97.2                                          \\

\bottomrule[1.5px]
\end{tabular}
\caption{Influence of bathsize on $D^{2}$-VPR.}
\label{batchsize}
\vspace{-2mm}
\end{table}

\begin{table*}[t]
\setlength{\extrarowheight}{0pt}
\setlength{\aboverulesep}{0pt}
\setlength{\belowrulesep}{0pt}
\small
\centering
\setlength{\tabcolsep}{1.4pt}
\renewcommand{\arraystretch}{1.2}
{

\begin{tabular}{@{}l||c||cc||cc||cc||cc||cc||cc}
\toprule[1.5px]
\multirow{2}{*}{Method} 
& \multirow{2}{*}{\scriptsize Dinov2}
& \multicolumn{2}{c||}{Pitts30k} 
& \multicolumn{2}{c||}{MSLS-val} 
& \multicolumn{2}{c||}{SPED} 
& \multicolumn{2}{c||}{Pitts250k} 
& \multicolumn{2}{c||}{ArmsterTime} 
& \multicolumn{2}{c}{Nordland} \\
\cline{3-14}
& & R@1 & R@5& R@1 & R@5& R@1 & R@5  & R@1 & R@5& R@1 & R@5& R@1 & R@5  \\
\hline

CricaVPR~$_{{\text{encoder, batchszie=1}}}$ 
&Base  & \textbf{91.8 } &  95.8 & 89.1
&  95.1 &   \textbf{87.6} &  \textbf{93.1}
&  94.6 & 98.3  &  \textbf{49.0}
&  69.3& \textbf{80.5} &  \textbf{90.7} \\

$D^{2}$-VPR~$_{{\text{encoder, batchszie=1}}}$        
& Small & \textbf{91.8} &  \textbf{96.2} & \textbf{91.1}
&  \textbf{96.1} &83.2  &91.4
& \textbf{94.9} &  \textbf{98.4} &48.7
& \textbf{69.6} &77.6 & 88.5 \\

\hline

SuperVLAD~$_{{\text{no encoder}}}$ 
&Base  & \textbf{92.6} &  \textbf{96.2} & 90.7 
&  \textbf{95.5} &   \textbf{89.1} &  \textbf{94.4}
&  \textbf{95.2} & \textbf{98.4}  &  \textbf{50.6}
&  \textbf{73.4}& 74.6 & 85.7 \\

$D^{2}$-VPR~$_{{\text{no encoder}}}$        
& Small & 91.6 & 95.8 &  \textbf{90.9}
&\textbf{95.5} &86.2  & 93.2
& 94.6&  \textbf{98.4} &49.1
& 70.3 & \textbf{78.0} &  \textbf{89.1} \\

\bottomrule[1.5px]
\end{tabular}}
\caption{Performance comparison with methods that use cross-image encoders (CricaVPR and SuperVLAD) when the cross-image encoder is not employed. Here, $D^{2}$-VPR and CricaVPR do not use PCA.}
\label{encoder}
\end{table*}

Our method follows the idea of CricaVPR~\cite{crica} and SuperVLAD~\cite{lu2024supervlad} and further improves performance through a cross-image encoder. This requires setting an appropriate batch size during the inference stage (8 for Amstertime, Pitts30k and Pitts250k. 3 for SPED. 128 for Nordland and MSLS-val.). We analyze how its performance changes with the inference batch size in Table \ref{batchsize} on msls-val. 

In Table \ref{encoder}, we further compare our method against other SOTA methods using the cross-image encoder, which indicates that when none of the methods use the cross-image encoder (when the batch size is set to 1, no cross-image feature interaction occurs), our VPR method can still achieve competitive performance. 

\subsection{B. Influence of PCA}

\begin{table*}[h]
\setlength{\extrarowheight}{0pt}
\setlength{\aboverulesep}{0pt}
\setlength{\belowrulesep}{0pt}
\small
\centering
\setlength{\tabcolsep}{2pt}
\renewcommand{\arraystretch}{1.2}

\begin{tabular}{@{}l||c||cc||cc||cc||cc||cc||cc}
\toprule[1.5px]
\multirow{2}{*}{Method} 
& \multirow{2}{*}{\scriptsize Dim.}

& \multicolumn{2}{c||}{Pitts30k} 
& \multicolumn{2}{c||}{MSLS-val}

& \multicolumn{2}{c||}{SPED} 
& \multicolumn{2}{c||}{Pitts250k}

& \multicolumn{2}{c||}{Amstertime} 
& \multicolumn{2}{c}{Nordland} \\

\cline{3-14}
&  & R@1 & R@5 & R@1 & R@5 & R@1 & R@5 & R@1 & R@5 & R@1 & R@5 & R@1 & R@5 \\
\hline

$D^{2}$-VPR~$_{{\text{encoder}}}$      
& 10752
& \textbf{94.9} & \textbf{97.3}
& \textbf{91.6} & \textbf{96.5}
& \textbf{91.3} & \textbf{96.2}
& \textbf{97.8} & \textbf{99.4}
& \textbf{62.9} & 80.7
& \textbf{87.2} & \textbf{94.5} \\

$D^{2}$-VPR~$_{{\text{encoder}}}$      +PCA 
& 4096
& \textbf{94.9} & \textbf{97.3}
& \textbf{91.6} & 96.1
& 90.9 & 96.0
& \textbf{97.8} & \textbf{99.4}
& \textbf{62.9} & \textbf{80.8}
& 86.6 & 94.1 \\
\hline

$\text{BoQ}$
& 16384 
& \textbf{92.1} & \textbf{95.8}
& \textbf{87.8} & \textbf{92.3 }
& \textbf{84.2} & \textbf{93.6}
& \textbf{94.7} & \textbf{98.2} 
& \textbf{43.5} & \textbf{62.3} 
& \textbf{79.5} & \textbf{88.9}
\\

$\text{BoQ}$ +PCA
& 4096 
& 92.0 & 95.6 
& 86.6 & \textbf{92.3 }
& 82.7 & 91.3
& 94.4 & 97.9 
& 42.0 & 60.5 
& 78.9 & 88.5
\\

\bottomrule[1.5px]
\end{tabular}
\caption{Comparison of feature dimensions vs. performance (R@1 and R@5) across six datasets.}
\label{pca_all}
\end{table*}

Building on the approach of ~\cite{crica}, we leverage PCA for dimensionality reduction. Tables \ref{pca_all} systematically evaluate the impact of PCA on feature compactness and retrieval performance across multiple datasets (Pitts30k, MSLS-val, etc.).

\subsection{C. Detials of $D^{2}$-VPR}

\begin{table}[h]
\centering
\setlength{\tabcolsep}{3mm}
\renewcommand\arraystretch{1.2}

\begin{tabular}{l||ccc} 
\toprule[1.5px]
Component   & Param. (M) \\
\hline

Adapter                       &  2.30                                        \\ 

Backbone          & 22.16                                    \\
Distill Recovery Module           & 1.17                                        \\

Deformable Aggregator           & 1.58                                      \\
Cross-image Encode           & 11.03                                       \\
All           & 38.24                                        \\

\bottomrule[1.5px]
\end{tabular}
\caption{Analysis of model parameter quantity.}
\label{backbone}
\vspace{-2mm}
\end{table}

Table~\ref{backbone} provides a breakdown of the parameter distribution across the components of the proposed model. The backbone (DINOv2-small) contains 22.16M parameters, accounting for the largest portion of the total. The adapter modules follow with 2.30M parameters. The distill recovery module includes 1.17M parameters. The deformable aggregator comprises 1.58M parameters, and the cross-image encoder contains 11.03M parameters. Summing all components, the complete model has 38.24M parameters.

\subsection{D. Attention Visualization}
\begin{figure*}[h]
    \centering
    \includegraphics[width=1\linewidth]{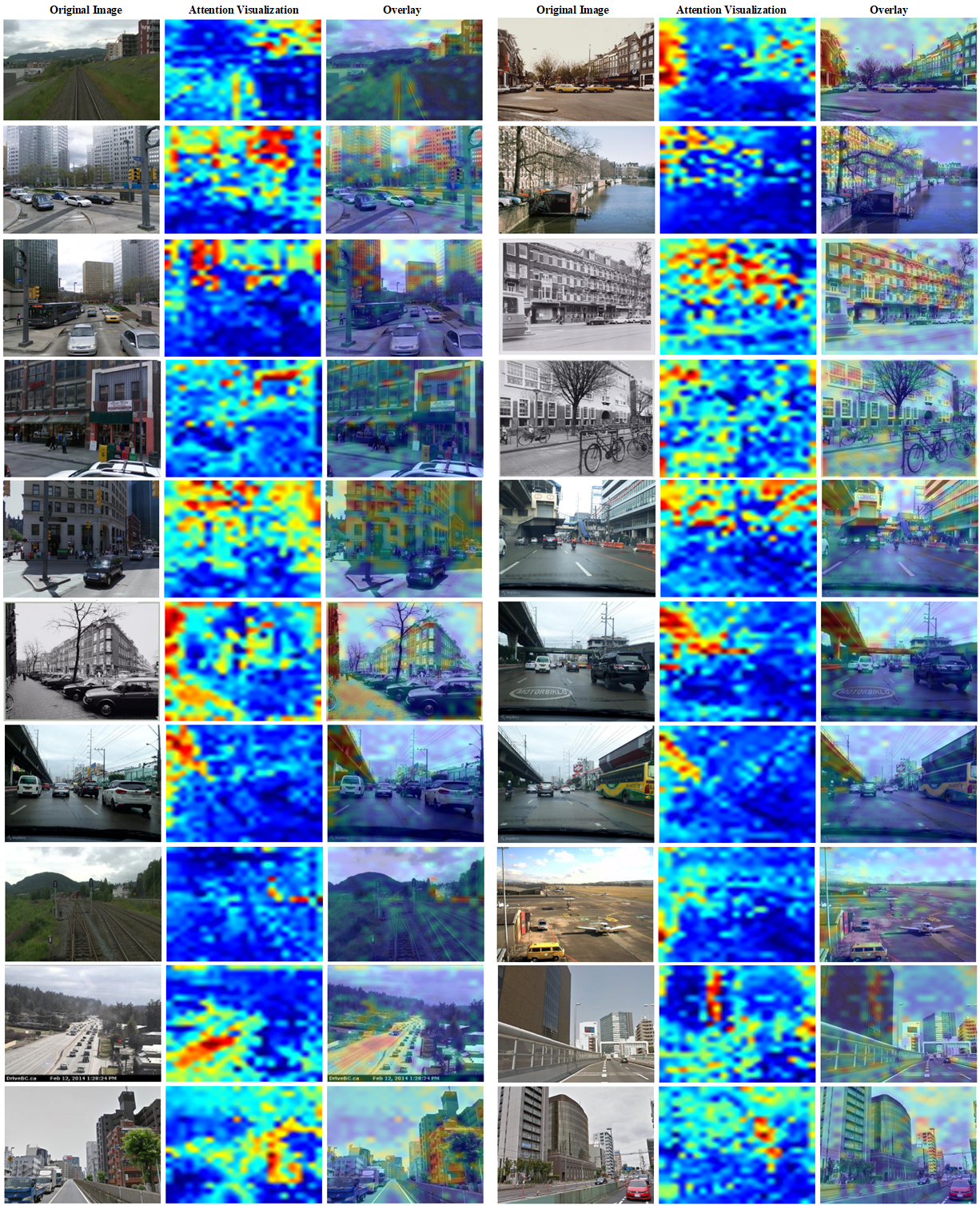}
    \caption{Qualitative attention map of our model. The images are sourced from Pitts30k, Pitts250k, MSLS, Nordland, SPED, and AmsterTime. Their content ranges from natural scenes to urban environments, and they include both black-and-white and color images.}
\label{attention}
\vspace{-0.3cm}
\end{figure*}

In Figure \ref{attention}, we present the attention visualizations of $D^{2}$-VPR for images from Pitts30k~\cite{pitts}, Pitts250k~\cite{pitts}, MSLS~\cite{msls}, Nordland~\cite{nordland}, SPED~\cite{SPED}, and AmsterTime~\cite{amstertime}. It is observed that $D^{2}$-VPR effectively focuses on objects in key regions of VPR that remain stable over time, such as buildings and bridges. Meanwhile, it maintains low attention on irrelevant regions and objects. For instance: in the left example of the first row, the sky; in the right example of the first row, cars and road surfaces; in the left example of the second row, cars; in the right example of the second row, cars, tree trunks, and lake surfaces; in both left and right examples of the third row, cars; in the left example of the fourth row, cars; in the right example of the fourth row, tree trunks and bicycles; in the fifth, sixth, and seventh rows, cars and road surfaces; in the left example of the eighth row, trees; in the right example of the eighth row, airplanes; and in the ninth and tenth rows, cars.

\subsection{E. Details of the Datasets}

\noindent\textbf{Pitts30k and Pitts250k}~\cite{pitts} are large-scale image retrieval datasets commonly used for evaluating VPR systems. Both datasets are constructed from Google Street View panoramas captured in the city of Pittsburgh and contain perspective-view images associated with GPS coordinates. Pitts30k is a smaller subset typically adopted for training and rapid evaluation, whereas Pitts250k provides a much larger and more challenging benchmark. These datasets serve as standard benchmarks for assessing the robustness and generalization capability of VPR methods in real-world urban environments.

\noindent\textbf{MSLS}~\cite{msls} is a large-scale dataset for evaluating visual place recognition and long-term localization under real-world conditions. It contains over 1.6 million images collected from 30 cities across six continents, covering diverse seasons, weather conditions, times of day, and camera devices. Images are organized into sequences with GPS and compass data, enabling robust benchmarking for sequence-based and single-image VPR methods. MSLS is particularly challenging due to its extreme variations in viewpoint and appearance, making it a key benchmark for testing model generalization in global-scale applications.

\noindent\textbf{Nordland}~\cite{nordland} is a VPR dataset derived from a train journey repeatedly recorded along the same railway route in Norway during different seasons. The recordings were captured using a front-facing camera mounted on the train, producing multiple time-aligned video sequences. The dataset is challenging due to the substantial appearance changes caused by seasonal variation, while viewpoint and motion remain nearly identical. It is widely used as a benchmark for assessing the robustness of VPR models under significant environmental and appearance changes.

\noindent\textbf{SPED (test)}~\cite{SPED} is a curated subset of the original SPED dataset, introduced in the VPR-Bench framework for standardized evaluation of VPR algorithms. It comprises images captured from fixed-position outdoor surveillance cameras under extreme appearance variations such as day/night, seasonal, and weather changes. The dataset emphasizes long-term appearance change while maintaining fixed viewpoints, making it ideal for testing a model’s robustness to illumination and temporal variation in place recognition tasks.

\noindent\textbf{AmsterTime}~\cite{amstertime} is a recently introduced dataset focused on evaluating long-term VPR under extreme appearance changes over years. Collected from public webcams in Amsterdam, it contains time-lapse images spanning over a decade, covering changes in lighting, weather, seasons, and even structural modifications in the environment. Unlike many other datasets, AmsterTime highlights the challenges of long-term autonomy, making it valuable for benchmarking models that aim to operate reliably over extended periods in the same environment.

\newpage

\bibliography{aaai2026}

@inproceedings{sela,
  title={Towards Seamless Adaptation of Pre-trained Models for Visual Place Recognition},
  author={Lu, Feng and Zhang, Lijun and Lan, Xiangyuan and Dong, Shuting and Wang, Yaowei and Yuan, Chun},
  booktitle={ICLR},
  year={2024}
}

@inproceedings{crica,
  title={CricaVPR: Cross-image Correlation-aware Representation Learning for Visual Place Recognition},
  author={Lu, Feng and Lan, Xiangyuan and Zhang, Lijun and Jiang, Dongmei and Wang, Yaowei and Yuan, Chun},
  booktitle={Proceedings of the IEEE/CVF Conference on Computer Vision and Pattern Recognition},

  year={2024}
}

@InProceedings{salad,
    author    = {Izquierdo, Sergio and Civera, Javier},
    title     = {Optimal Transport Aggregation for Visual Place Recognition},
    booktitle = {Proceedings of the IEEE/CVF Conference on Computer Vision and Pattern Recognition},
    
    year      = {2024},
}

@inproceedings{ali2024boq,
  title={BoQ: A Place is Worth a Bag of Learnable Queries},
  author={Ali-bey, Amar and Chaib-draa, Brahim and Gigu{\`e}re, Philippe and},
  booktitle={Proceedings of the IEEE/CVF Conference on Computer Vision and Pattern Recognition},
  pages={17794--17803},
  year={2024}
}

@inproceedings{lu2024supervlad,
  title={SuperVLAD: Compact and Robust Image Descriptors for Visual Place Recognition},
  author={Lu, Feng and Zhang, Xinyao and Ye, Canming and Dong, Shuting and Zhang, Lijun and Lan, Xiangyuan and Yuan, Chun},
  booktitle={Advances in Neural Information Processing Systems},
  volume={37},
  pages={5789--5816},
  year={2024}
}

@ARTICLE{EDTformer,
  author={Jin, Tong and Lu, Feng and Hu, Shuyu and Yuan, Chun and Liu, Yunpeng},
  journal={IEEE Transactions on Circuits and Systems for Video Technology}, 
  title={EDTformer: An Efficient Decoder Transformer for Visual Place Recognition}, 
  year={2025},
  doi={10.1109/TCSVT.2025.3559084}}

@inproceedings{wang2025focus,
  title={Focus on Local: Finding Reliable Discriminative Regions for Visual Place Recognition},
  author={Wang, Changwei and Chen, Shunpeng and Song, Yukun and Xu, Rongtao and Zhang, Zherui and Zhang, Jiguang and Yang, Haoran and Zhang, Yu and Fu, Kexue and Du, Shide and others},
  booktitle={Proceedings of the AAAI Conference on Artificial Intelligence},
  volume={39},
  number={7},
  pages={7536--7544},
  year={2025}
}

@inproceedings{pitts,
  title={Visual place recognition with repetitive structures},
  author={Torii, Akihiko and Sivic, Josef and Pajdla, Tomas and Okutomi, Masatoshi},
  booktitle={Proceedings of the IEEE conference on computer vision and pattern recognition},
  pages={883--890},
  year={2013}
}

@inproceedings{msls,
  title={Mapillary street-level sequences: A dataset for lifelong place recognition},
  author={Warburg, Frederik and Hauberg, Soren and Lopez-Antequera, Manuel and Gargallo, Pau and Kuang, Yubin and Civera, Javier},
  booktitle={Proceedings of the IEEE/CVF conference on computer vision and pattern recognition},
  pages={2626--2635},
  year={2020}
}

@inproceedings{amstertime,
  title={Amstertime: A visual place recognition benchmark dataset for severe domain shift},
  author={Yildiz, Burak and Khademi, Seyran and Siebes, Ronald Maria and Van Gemert, Jan},
  booktitle={2022 26th International Conference on Pattern Recognition (ICPR)},
  pages={2749--2755},
  year={2022},
  organization={IEEE}
}

@article{nordland,
  title={Are we there yet? Challenging SeqSLAM on a 3000 km journey across all four seasons},
  author={S{\"u}nderhauf, Niko and Neubert, Peer and Protzel, Peter},
  booktitle={Proc. of workshop on long-term autonomy, IEEE international conference on robotics and automation (ICRA)},
  pages={2013},
  year={2013}
}

@article{SPED,
  title={Vpr-bench: An open-source visual place recognition evaluation framework with quantifiable viewpoint and appearance change},
  author={Zaffar, Mubariz and Garg, Sourav and Milford, Michael and Kooij, Julian and Flynn, David and McDonald-Maier, Klaus and Ehsan, Shoaib},
  journal={International Journal of Computer Vision},
  pages={1--39},
  year={2021},
  publisher={Springer}
}

@inproceedings{wang2019multi,
title={Multi-Similarity Loss with General Pair Weighting for Deep Metric Learning},
author={Wang, Xun and Han, Xintong and Huang, Weilin and Dong, Dengke and Scott, Matthew R},
booktitle={Proceedings of the IEEE Conference on Computer Vision and Pattern Recognition},
pages={5022--5030},
year={2019}
}

@article{nie2025epsa,
  title={EPSA-VPR: A lightweight visual place recognition method with an Efficient Patch Saliency-weighted Aggregator},
  author={Nie, Jiwei and Zh{\`a}o, Qǐx{\i}̄ and Xue, Dingyu and Pan, Feng and Liu, Wei},
  journal={Journal of Visual Communication and Image Representation},
  pages={104440},
  year={2025},
  publisher={Elsevier}
}

@article{xu2024clusvpr,
  title={ClusVPR: Efficient Visual Place Recognition with Clustering-based Weighted Transformer},
  author={Xu, Yifan and Shamsolmoali, Pourya and Zareapoor, Masoume and Yang, Jie},
  journal={IEEE Transactions on Artificial Intelligence},
  year={2024},
  publisher={IEEE}
}

@article{nie2024efficient,
  title={Efficient saliency encoding for visual place recognition: Introducing the lightweight pooling-centric saliency-aware VPR method},
  author={Nie, Jiwei and Xue, Dingyu and Pan, Feng and Ning, Zuotao and Liu, Wei and Hu, Jun and Cheng, Shuai},
  journal={IEEE Robotics and Automation Letters},
  volume={9},
  number={7},
  pages={6035--6042},
  year={2024},
  publisher={IEEE}
}

@article{SURF,
  title={Speeded-up robust features (SURF)},
  author={Bay, H. and Ess, A. and Tuytelaars, T. and Gool, L. V.},
  journal={Computer vision and image understanding},
  volume={110},
  number={3},
  pages={346--359},
  year={2008},
  publisher={Elsevier}
}

@inproceedings{netvlad,
  title={NetVLAD: CNN architecture for weakly supervised place recognition},
  author={Arandjelovic, Relja and Gronat, Petr and Torii, Akihiko and Pajdla, Tomas and Sivic, Josef},
  booktitle={CVPR},
  pages={5297--5307},
  year={2016}
}

@inproceedings{patchvlad,
  title={Patch-netvlad: Multi-scale fusion of locally-global descriptors for place recognition},
  author={Hausler, Stephen and Garg, Sourav and Xu, Ming and Milford, Michael and Fischer, Tobias},
  booktitle={CVPR},
  pages={14141--14152},
  year={2021}
}

@article{oquab2023dinov2,
  title={Dinov2: Learning robust visual features without supervision},
  author={Oquab, Maxime and Darcet, Timoth{\'e}e and Moutakanni, Th{\'e}o and Vo, Huy and Szafraniec, Marc and Khalidov, Vasil and Fernandez, Pierre and Haziza, Daniel and Massa, Francisco and El-Nouby, Alaaeldin and others},
  journal={arXiv preprint arXiv:2304.07193},
  year={2023}
}

@article{huang2024dino,
  title={DINO-Mix enhancing visual place recognition with foundational vision model and feature mixing},
  author={Huang, Gaoshuang and Zhou, Yang and Hu, Xiaofei and Zhang, Chenglong and Zhao, Luying and Gan, Wenjian},
  journal={Scientific Reports},
  volume={14},
  number={1},
  pages={22100},
  year={2024},
  publisher={Nature Publishing Group UK London}
}

@article{tzachor2024effovpr,
  title={Effovpr: Effective foundation model utilization for visual place recognition},
  author={Tzachor, Issar and Lerner, Boaz and Levy, Matan and Green, Michael and Shalev, Tal Berkovitz and Habib, Gavriel and Samuel, Dvir and Zailer, Noam Korngut and Shimshi, Or and Darshan, Nir and others},
  journal={arXiv preprint arXiv:2405.18065},
  year={2024}
}

@article{wan2025scicevpr,
  title={SciceVPR: Stable Cross-Image Correlation Enhanced Model for Visual Place Recognition},
  author={Wan, Shanshan and Wei, Yingmei and Kang, Lai and Shen, Tianrui and Wang, Haixuan and Yang, Yee-Hong},
  journal={arXiv preprint arXiv:2502.20676},
  year={2025}
}

@inproceedings{mixvpr,
  title={{MixVPR:} Feature Mixing for Visual Place Recognition},
  author={Ali-bey, Amar and Chaib-draa, Brahim and Gigu{\`e}re, Philippe},
  booktitle={WACV},
  pages={2998--3007},
  year={2023}
}

@article{keetha2023anyloc,
  title={Anyloc: Towards universal visual place recognition},
  author={Keetha, Nikhil and Mishra, Avneesh and Karhade, Jay and Jatavallabhula, Krishna Murthy and Scherer, Sebastian and Krishna, Madhava and Garg, Sourav},
  journal={IEEE Robotics and Automation Letters},
  year={2023},
  publisher={IEEE}
}

@inproceedings{chen2017deep,
  title={Deep learning features at scale for visual place recognition},
  author={Chen, Zetao and Jacobson, Adam and S{\"u}nderhauf, Niko and Upcroft, Ben and Liu, Lingqiao and Shen, Chunhua and Reid, Ian and Milford, Michael},
  booktitle={2017 IEEE international conference on robotics and automation (ICRA)},
  pages={3223--3230},
  year={2017},
  organization={IEEE}
}

@article{lowry2015visual,
  title={Visual place recognition: A survey},
  author={Lowry, Stephanie and S{\"u}nderhauf, Niko and Newman, Paul and Leonard, John J and Cox, David and Corke, Peter and Milford, Michael J},
  journal={ieee transactions on robotics},
  volume={32},
  number={1},
  pages={1--19},
  year={2015},
  publisher={IEEE}
}

@article{ventura2014global,
  title={Global localization from monocular slam on a mobile phone},
  author={Ventura, Jonathan and Arth, Clemens and Reitmayr, Gerhard and Schmalstieg, Dieter},
  journal={IEEE transactions on visualization and computer graphics},
  volume={20},
  number={4},
  pages={531--539},
  year={2014},
  publisher={IEEE}
}

@inproceedings{torii201524,
  title={24/7 place recognition by view synthesis},
  author={Torii, Akihiko and Arandjelovic, Relja and Sivic, Josef and Okutomi, Masatoshi and Pajdla, Tomas},
  booktitle={Proceedings of the IEEE conference on computer vision and pattern recognition},
  pages={1808--1817},
  year={2015}
}

@article{ali2022gsv,
  title={Gsv-cities: Toward appropriate supervised visual place recognition},
  author={Ali-bey, Amar and Chaib-draa, Brahim and Giguere, Philippe},
  journal={Neurocomputing},
  volume={513},
  pages={194--203},
  year={2022},
  publisher={Elsevier}
}

@inproceedings{sarlin2019coarse,
  title     = {From Coarse to Fine: Robust Hierarchical Localization at Large Scale},
  author    = {Paul-Edouard Sarlin and
               Cesar Cadena and
               Roland Siegwart and
               Marcin Dymczyk},
  booktitle = {CVPR},
  year      = {2019}
}

@inproceedings{lou2025overlock,
  title={OverLoCK: An Overview-first-Look-Closely-next ConvNet with Context-Mixing Dynamic Kernels},
  author={Lou, Meng and Yu, Yizhou},
  booktitle={Proceedings of the Computer Vision and Pattern Recognition Conference},
  pages={128--138},
  year={2025}
}

@article{GILBERT2007677,
title = {Brain States: Top-Down Influences in Sensory Processing},
journal = {Neuron},
volume = {54},
number = {5},
pages = {677-696},
year = {2007},
issn = {0896-6273},
doi = {https://doi.org/10.1016/j.neuron.2007.05.019},
url = {https://www.sciencedirect.com/science/article/pii/S0896627307003765},
author = {Charles D. Gilbert and Mariano Sigman}
}

@article{gou2021knowledge,
  title={Knowledge distillation: A survey},
  author={Gou, Jianping and Yu, Baosheng and Maybank, Stephen J and Tao, Dacheng},
  journal={International journal of computer vision},
  volume={129},
  number={6},
  pages={1789--1819},
  year={2021},
  publisher={Springer}
}

@article{radenovic2018fine,
  title={Fine-tuning CNN image retrieval with no human annotation},
  author={Radenovi{\'c}, Filip and Tolias, Giorgos and Chum, Ond{\v{r}}ej},
  journal={IEEE transactions on pattern analysis and machine intelligence},
  volume={41},
  number={7},
  pages={1655--1668},
  year={2018},
  publisher={IEEE}
}

@ARTICLE{10475537,
  author={Miao, Jinyu and Jiang, Kun and Wen, Tuopu and Wang, Yunlong and Jia, Peijin and Wijaya, Benny and Zhao, Xuhe and Cheng, Qian and Xiao, Zhongyang and Huang, Jin and Zhong, Zhihua and Yang, Diange},
  journal={IEEE Transactions on Intelligent Vehicles}, 
  title={A Survey on Monocular Re-Localization: From the Perspective of Scene Map Representation}, 
  year={2024},
  volume={},
  number={},
  pages={1-33},
  doi={10.1109/TIV.2024.3378716}}

@misc{grainge2025tetravpr,
      title={TeTRA-VPR: A Ternary Transformer Approach for Compact Visual Place Recognition},
      author={Oliver Grainge and Michael Milford and Indu Bodala and Sarvapali D. Ramchurn and Shoaib Ehsan},
      year={2025},
      eprint={2503.02511},
      archivePrefix={arXiv},
      primaryClass={cs.CV},
      url={https://arxiv.org/abs/2503.02511},
}

@article{kingma2014adam,
  title={Adam: A method for stochastic optimization},
  author={Kingma, Diederik P and Ba, Jimmy},
  journal={arXiv preprint arXiv:1412.6980},
  year={2014}
}

@article{loshchilov2017fixing,
  title={Fixing weight decay regularization in adam},
  author={Loshchilov, Ilya and Hutter, Frank and others},
  journal={arXiv preprint arXiv:1711.05101},
  volume={5},
  number={5},
  pages={5},
  year={2017}
}

@InProceedings{Berton_2022_CVPR,
    author    = {Berton, Gabriele and Masone, Carlo and Caputo, Barbara},
    title     = {Rethinking Visual Geo-Localization for Large-Scale Applications},
    booktitle = {Proceedings of the IEEE/CVF Conference on Computer Vision and Pattern Recognition (CVPR)},
    month     = {June},
    year      = {2022},
    pages     = {4878-4888}
}

@InProceedings{Ali-bey_2023_WACV,
    author    = {Ali-bey, Amar and Chaib-draa, Brahim and Gigu\`ere, Philippe},
    title     = {MixVPR: Feature Mixing for Visual Place Recognition},
    booktitle = {Proceedings of the IEEE/CVF Winter Conference on Applications of Computer Vision (WACV)},
    month     = {January},
    year      = {2023},
    pages     = {2998-3007}
}

@InProceedings{Berton_2023_ICCV,
    author    = {Berton, Gabriele and Trivigno, Gabriele and Caputo, Barbara and Masone, Carlo},
    title     = {EigenPlaces: Training Viewpoint Robust Models for Visual Place Recognition},
    booktitle = {Proceedings of the IEEE/CVF International Conference on Computer Vision (ICCV)},
    month     = {October},
    year      = {2023},
    pages     = {11080-11090}
}

\end{document}